\documentclass[letterpaper, 10 pt, conference]{ieeeconf}  

\IEEEoverridecommandlockouts                              

\usepackage{graphicx}
\usepackage{todonotes}
\usepackage[most]{tcolorbox}

\usepackage{amsthm}
\newtheorem{definition}{Definition}

\title{\LARGE \bf
Towards Contrastive Explanations for Comparing the Ethics of Plans
}

\author{Benjamin Krarup$^{1}$, Senka Krivic$^{1}$, Felix Lindner$^{2}$, and Derek Long$^{1}$
\thanks{*This work was supported by AFOSR and THuMP projects.}
\thanks{$^{1}$Authors are with the Department of Informatics,
        King's College London, UK
        {\tt\small \textit{firstname.lastname}@kcl.ac.uk}}%
\thanks{$^{2}$Felix Lindner is with the Institute of Artificial Intelligence, Ulm University, Germany
        {\tt\small felix.lindner@uni-ulm.de}}%
}

\renewcommand{\baselinestretch}{0.97}

\begin{document}

\maketitle
\thispagestyle{empty}
\pagestyle{empty}

\begin{abstract}
The development of robotics and AI agents has enabled their wider usage in human surroundings. 
AI agents are more trusted to make increasingly important decisions with potentially critical outcomes.
It is essential to consider the ethical consequences of the decisions made by these systems. In this paper, we present how contrastive explanations can be used for comparing the ethics of plans. We build upon an existing ethical framework to allow users to make suggestions about plans and receive contrastive explanations.
\end{abstract}

\section{INTRODUCTION}

AI Planning systems, referred to as planners, are used in a variety of complex domains to create a sequence of actions known as a plan to achieve a set of goals from an initial state. 
We are interested in models where actions are deterministic, durationless, and can be performed one at a time. We also assume a known initial state and goal.
Traditionally, ethical principles of single decisions are evaluated~\cite{dri10}.
In the context of AI Planning this means analysing a massive number of isolated decisions that may not make sense without the context in which they are being made.
Therefore, it is preferable to evaluate the ethical contents of a plan as a whole.
Lindner et al.~\cite{moralplanning} describe an approach to judging the ethical outcomes of an entire plan. 
We build upon their work by providing contrastive explanations~\cite{Cashmore_icapsxai2019, ben19} to allow users to understand the ethics of plans.

\begin{figure}[!t]
    \centering
    \includegraphics[width=0.7\columnwidth]{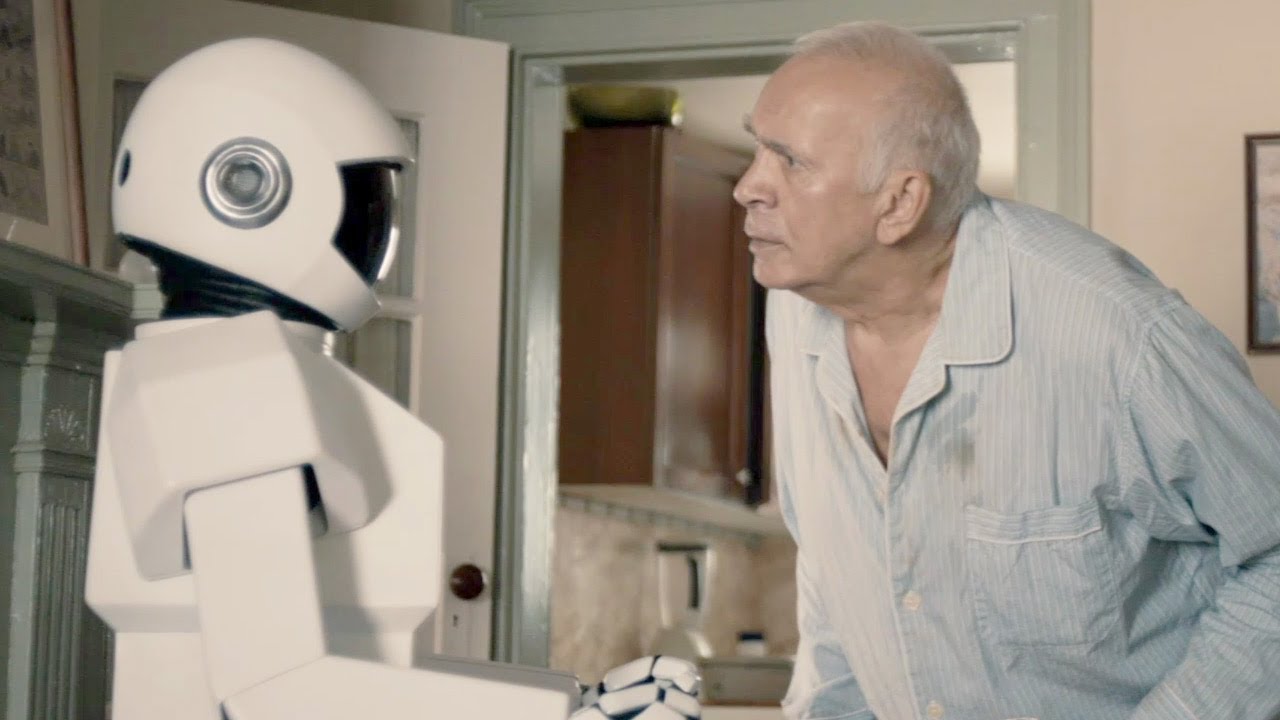}
    \caption{Motivating scene from the movie "Robot \& Frank". Snapshot is taken from the same movie. }
    \label{fig:motiv}
\end{figure}

We use as a running example the following ethical dilemma. A robot has been charged with the care of an elderly gentleman named Frank (Fig.~\ref{fig:motiv}). Frank has grown quite fond of the robot. The robot wants to motivate Frank so that he will do some exercise and keep healthy. The robot has two choices, it can either beg Frank to exercise, or it can lie. The robot can deceive Frank, exploiting the affection that Frank has for him, by telling Frank that he will be decommissioned if it cannot keep Frank healthy. 
A moderator is examining the robot's behaviour and
trying to understand the ethics of the plans produced, to ensure the plan is adhering to the moderator's moral principles.
Frank being healthy, has a positive utility and a positive effect on Frank, while Frank being unhealthy has a negative effect on him. The action of begging Frank is intrinsically neutral, while the action of lying to Frank is intrinsically bad.

Evaluating the ethics of individual plans is necessary for the integration of AI systems in everyday use.
However, it is not enough to pass judgement on automated plans. 
Users should be able to make suggestions, 
when they suspect the system could behave more ethically,
to produce alternative plans and compare the moralities of each.


This can be done through \textit{contrastive explanations}~\cite{mil18},
which focus on explaining the difference between a factual event \textit{A} and a contrasting event \textit{B}.
To produce these explanations, one must reason about the hypothetical alternative \textit{B}, which likely means constructing an alternative plan where \textit{B} is included rather than \textit{A}. The original model is constrained to produce a hypothetical planning model (HModel). The solution to the HModel is the hypothetical plan (HPlan) that contains the contrast case expected by the user.

The plans are then tested under the user's chosen ethical principle, and explanations are generated. A contrastive explanation is generated to show the difference in the moral permissibility of the plan and HPlan~\cite{lip90,lew86}.

Our proposed approach can be implemented \textit{as a service} -- i.e., as a wrapper around an existing planning system that takes as input the current planning problem and domain model, the current plan, and the user's suggestion and ethical principle. 
It can invoke the existing planning system on hypothetical problems which satisfy the user's suggestions. 
This approach allows users to get explanations constructed from their own trusted planner and model. 
This can alleviate concerns over the ethical behaviour of external systems.

\section{MORAL PERMISSIBILITY OF PLANS}

Researchers in ethics have proposed various ethical theories offering different ethical principles according to which the moral permissibility of actions can be judged: {\it consequentialist theories} judge actions based on their consequences, {\it deontological theories} judge actions based on their intrinsic value, by their intentions or means, and {\it hybrid theories}, such as the principle of double effect, combine consequentialism and deontology. Lindner et al.~\cite{moralplanning} present an approach to checking the moral permissibility of action plans according to a collection of moral principles~\footnote{We provide explanations for all principles discussed in Lindnder et al.~\cite{moralplanning}}. The approach is based on the idea that moral permissibility checking can be reduced to model checking. Formulae that can be evaluated over action plans capture the meaning of the respective ethical theory. For example, the formula $\phi = \bigwedge_{a \in Pi} \lnot Bad(a)$ captures the deontological rule that none of the actions in a plan may be intrinsically bad. We use prime implicate generation methods to find sufficient and necessary reasons for $\phi$ to become true or false \cite{LindnerMoellney2019}. In the Robot \& Frank case, lying is intrinsically bad. Thus, $\phi$ is false for every plan that involves lying. The prime implicate $Bad(lying)$ is a sufficient and necessary reason for $\lnot \phi$ and hence explains the moral impermissibility of such a plan.
We use a framework for explanation construction based on this approach which we refer to as the \textit{Ethical Explanation Generator}.

\section{CONTRASTIVE EXPLANATIONS}


A plan produced by an AI agent may not adhere to the ethical principles of a human moderator. 
The moderator may have insight on how to improve the ethical outcome of the plan, 
but not have the ability to evaluate their prediction.
To this end, we allow a human-in-the-loop to suggest improvements in the plan.
The AI agent can plan accordingly for these suggestions and explain to the user how the original and new plans differ ethically.
Contrastive explanations are naturally useful for comparison~\cite{hes88}.
Therefore, we believe contrastive explanations are instrumental in comparing the ethical outcomes of plans, 
allowing users to better understand the ethics of plans produced by planning systems.

\begin{definition}
An \textbf{explanation problem} is a tuple $E = \langle \Pi, \phi, \sigma, \epsilon \rangle$, in which $\Pi$ is a planning model, $\phi$ is the plan generated by the planner, $\sigma$ is the suggestion given by the user, and $\epsilon$ is the ethical principle. 
\end{definition}

The ability to reason about what would happen in the 
contrast cases suggested by the user is essential to compare the ethical outcomes of plans. 
We approach the problem in Definition~1 by generating plans for the 
contrast cases via \emph{compilations} which enforce the moderator's suggestion. 

A compilation of a planning instance defined by the planning model $\Pi$ is shown as $Compilation(\Pi, \sigma) = \Pi'$ where $\sigma$ is the suggestion~\footnote{Krarup et al.~\cite{ben19} provides a full formalisation of all the compilations used.}.
The compilation process derives constraints from the suggestion $\sigma$ which enforces the user's suggestion and produces a new planning model $\Pi'$.
We call $\Pi'$ the \textit{hypothetical model}, or HModel. 
However, $\Pi'$ can also be used as the input model so that the user can iteratively suggest improvements to a plan produced by some model, i.e
$Compilation(Compilation(\Pi, \sigma), \sigma')$.
After the HModel is formed, it is solved with the original planner to give the HPlan $\phi'$.
For each iteration of compilation, the HPlan is validated against the original model $\Pi$ to ensure that the plan suggested by the moderator is also applicable in the current situation.
An explanation for the permissibility of the both plans $\phi$ and $\phi'$ under $\epsilon$ is formed from the Ethical Explanation Generator.
A contrastive explanation highlights the differences between $\phi$ and $\phi'$, and therefore sheds light on the consequence of the user's suggestion. This process is visualised in Fig.~\ref{fig:overview}.

In our example, the robot may lie to Frank to motivate him to work out and stay healthy. The moderator might wonder about the ethical reasoning behind this decision. Before he passes judgement, the moderator wants to compare the ethical outcome of lying to Frank with the original plan of begging Frank to exercise. 
This forces the planner to perform the action beg Frank
to achieve the goal 
and restricts the model from performing the action lie to Frank. 
The HPlan produced from the HModel is now one where the robot begs Frank to exercise. 
The plan and HPlan are then inputted to the Ethical Explanation Generator, along with the ethical principle chosen by the moderator, in this case the deontological principle.
This produces two causal explanations \textit{\textit{Bad(lying to Frank) and $\lnot$Bad(begging Frank)}}.
We combine these explanations to produce a natural language contrastive explanation: \textit{"The original plan is impermissible because lying to Frank is bad, whereas the HPlan is permissible because begging Frank is not bad"}.

\section{CONCLUSION}
\begin{figure}[t!]
\centering
\includegraphics[width=\linewidth]{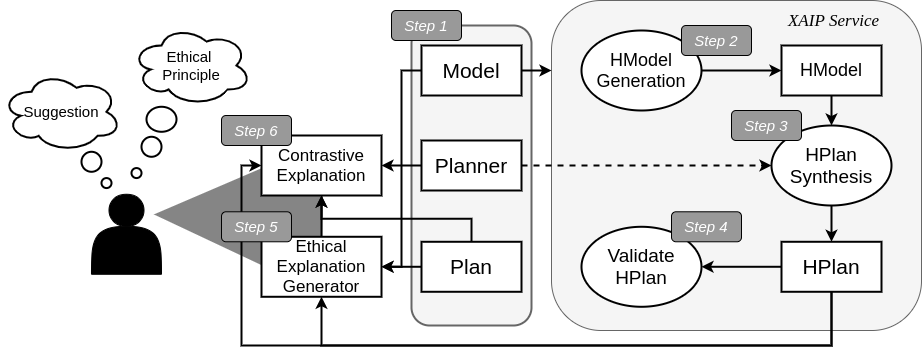}
\caption{Architecture diagram for the implemented system which provides contrastive explanations of the ethics of plans}
\label{fig:overview}
\end{figure}

We implemented a system for providing contrasting explanations for comparing the ethical outcomes of plans~\footnote{We have omitted the full details our implementation due to space constraints. However, the architecture diagram is shown in Fig.~\ref{fig:overview}.}.
In the future, we will conduct user studies to test our approach and verify our assumption that contrastive explanations are useful for users to understand the ethics of plan. 
Finally, we will evaluate whether this process helps to evolve the trust the user has in the system to produce ethical plans for autonomous robots.

\addtolength{\textheight}{-12cm}   





\bibliographystyle{IEEEtran}
\bibliography{bib}

\end{document}